\documentclass{article}
\pdfoutput=1
\usepackage[utf8]{inputenc}
\usepackage[a4paper, margin=1in]{geometry}
\usepackage[active]{srcltx}
\usepackage{babel}
\usepackage{verbatim}
\usepackage{mathtools}
\usepackage{bm}
\usepackage{amsmath}
\usepackage{amssymb}
\usepackage{microtype,comment}
\usepackage{graphicx}
\usepackage{subfigure}
\usepackage{booktabs} 
\usepackage{hyperref}
\usepackage{verbatim}
\usepackage{mathtools}
\usepackage{bm}
\usepackage{amsmath}
\usepackage{amssymb} 
\usepackage{amsthm}
\usepackage{babel}
\usepackage{amsthm}
\usepackage{dsfont}
\usepackage{array}
\usepackage{mathrsfs}
\usepackage{color}
\usepackage{natbib}
\usepackage{enumerate}
 \bibpunct[, ]{(}{)}{,}{a}{}{,}%
 \def\BIBand{and}%
 
\theoremstyle{plain} 
\newtheorem{lemma}{\textbf{Lemma}} 

\newtheorem{theorem}{\textbf{Theorem}}
\newtheorem{corollary}{\textbf{Corollary}}

\theoremstyle{definition}
\theoremstyle{remark}

\newtheorem{remark}{\textbf{Remark}}

\newcommand{\BB}{\mathbb{B}}
\newcommand{\RR}{\mathbb{R}}
\newcommand{\Scal}{\mathcal{S}}

\title{A Concentration Bound for TD(0) with Function Approximation}
\author{Siddharth Chandak\footnote{SC is with Stanford University, chandaks@stanford.edu}, Vivek S.\ Borkar\footnote{VSB is with Indian Institute of Technology Bombay, borkar.vs@gmail.com.}}

\date{July 2023}

\begin{document}
\maketitle
\begin{abstract}
We derive uniform all-time concentration bound of the type `for all $n \geq n_0$ for some $n_0$' for TD(0) with linear  function approximation. We work with online TD learning with samples from a single sample path of the underlying Markov chain. This makes our analysis significantly different from offline TD learning or TD learning with access to independent samples from the stationary distribution of the Markov chain. 
We treat TD(0) as a contractive stochastic approximation algorithm, with both martingale and Markov noises. Markov noise is handled using the Poisson equation and the lack of almost sure guarantees on boundedness of iterates is handled using the concept of relaxed concentration inequalities.
\end{abstract}


\section{Introduction}\label{sec-intro}

TD(0) is one of the most popular Reinforcement Learning (RL) algorithms for policy evaluation \citep{vanRoyTsitsiklis}. Given a fixed policy, the algorithm is an iterative method to obtain the value function for each state under the long-term discounted reward framework. To mitigate the issues of large state spaces, the value function is often approximated using a linear combination of feature vectors. This algorithm is referred to as TD(0) with linear function approximation. In this paper, we work with online TD(0) with a single sample path of the underlying Markov chain. Our goal in this paper is to obtain a concentration bound of the form 
\textit{from some time on} or more precisely, \textit{for all $n\geq n_0$ for a suitably chosen $n_0$} for this algorithm.

A bound of this form was published for TD(0) as a section in our paper titled ``Concentration of Contractive Stochastic Approximation and Reinforcement Learning'' \citep{chandak}. This paper established an all-time bound for contractive stochastic approximation with Markov noise and applied the bound to asynchronous Q-learning and TD(0). Although the main theorem and its application to asynchronous Q-learning are correct, TD(0) does not satisfy a key assumption for the main theorem\footnote{In \citet{chandak}, TD(0) does not satisfy assumption (6) which is required in the proof for Lemma 1 that shows almost sure boundedness of the iterates. The authors thank Zaiwei Chen for pointing this out.} and hence the theorem was incorrectly applied to TD(0). We remove the need for that assumption in this version, giving a completely different proof tailored to the TD(0) algorithm.

The previous paper required the iterates of the stochastic approximation iteration to be bounded by a constant with probability 1. This assumption is not known to be true for the iterates of online TD(0) with function approximation for a single sample path. In fact, a common method to alleviate this issue is to project the iterates back into a ball centered around the origin \citep{Bhandari, patil}. The key difficulty caused by the lack of this assumption is in applying martingale inequalities, which often require some restrictions on the increments of the martingale that are often not easy to verify. We do not modify the algorithm and instead adapt relaxed concentration inequalities \citep[Section 8]{Chung_Mart_ineq} for our problem. These bounds have an extra term given by the probability of increments going above a certain threshold \citep[Prop.\ 34]{tao}. While the proof in this paper is restricted to TD(0), the underlying idea of using relaxed concentration inequalities is broadly applicable to other algorithms that face similar challenges due to unboundedness.

\subsection{Related Works}
There has been growing interest in analyzing the finite-time performance of reinforcement learning (RL) algorithms. Existing results can broadly be categorized by the type of bounds they establish. The most extensive body of work concerns expectation or mean-square bounds (see, e.g., \cite{Chen, Chen2}). Another prominent line of research focuses on regret bounds, which characterize how the cumulative error grows over time—typically through almost-sure or expected regret guarantees (see, e.g.,  \cite{regret-azar, Chi_Jin, Yang, Lin_Yang}). A third class comprises high-probability or concentration bounds (see, e.g., \cite{Li-QL2, Qu, Mansour}. Our work falls within this category but differs from conventional analyses that establish high-probability guarantees only for sufficiently large time $n$.
In contrast, we derive uniform all-time bounds, i.e., bounds that hold for all $n\geq n_0$ with probability at least $1-\delta$.

Specifically for TD(0), moment bounds have been established in  \cite{Srikant}, \cite{Bhandari}, and  \cite{Chen2}. High probability bounds have been established under  various modifications of the TD(0) algorithm. These include uniform sampling from dataset \citep{Prashanth}, projection and tail averaging \citep{patil}, and oracle access to i.i.d.\ samples of state - action - next state triplets $(s,a,s')$ \citep{Zaiwei_alltime, Gugan2}.

One of us considered stochastic approximation involving contractive maps and martingale noise, and derived maximal concentration bounds for this class of algorithms \citep{Borkar-conc}. This covered, in particular, synchronous Q-learning for discounted cost and some related schemes. In \citet{chandak},  we extended this work to cover `Markov noise' \citep{Meerkov} in the stochastic approximation scheme, allowing us to give bounds for the asynchronous case. As mentioned before, this work assumed almost-sure boundedness of iterates, which is not satisfied by the TD(0) algorithm. We remove the need for this assumption in our current work. Other articles aiming at bounds of these forms, such as the one we provide, are found in \cite{chandak_lspe} for the LSPE algorithm, and \citet{Borkar0, Gugan, Kamal}  for abstract stochastic approximation schemes. A recent work \citep{Zaiwei_alltime} considers all-time bounds for iterates without almost-sure boundedness, but they only consider additive and multiplicative noise and not Markovian noise as considered in this paper. Their proof technique relies on Moreau envelopes and a bootstrapping technique, which differs significantly from our work.

\subsection{Outline and Notation}
The rest of the paper is structured as follows. Section \ref{sec:background} gives a background to the TD(0) algorithm, along with the required assumptions and the stochastic approximation formulation. Section \ref{sec:main-result} states the main result and provides some insights into the result. The result is proved in Section \ref{sec:proof}. A concluding section highlights some future directions. Appendix \ref{Appendix-main} states a martingale inequality used in our proof and Appendix \ref{Appendix-proofs} gives proofs for some technical lemmas which are used to prove the main theorem.

Throughout this work, $\|\cdot\|$ denotes the Euclidean norm on $\RR^d$, and $\langle\cdot,\cdot\rangle$ denotes the inner product in $\RR^d$. $\theta$ denotes the zero vector in $\RR^d$. The $\ell$\textsuperscript{th} component of a vector $x$ and a vector-valued function $h(\cdot)$ are denoted by $x(\ell)$ and $h^{\ell}(\cdot)$, respectively. 

\section{Background on TD(0)}\label{sec:background}
TD(0) is an algorithm for policy evaluation, i.e., for learning the performance of a fixed policy, and not for optimizing over policies. Hence a stationary policy is fixed a priori, giving us a time homogeneous uncontrolled Markov chain $\{Y_n\}$ over a finite state space $\mathcal{S}$. The transition probabilities are given by $p(\cdot|\cdot)$, where the dependence on the policy is suppressed. Assume that the chain is aperiodic irreducible with the stationary distribution $\pi=[\pi(1),...,\pi(S)], S=|\mathcal{S}|$. Let $D$ denote the $S\times S$ diagonal matrix whose $s$\textsuperscript{th} diagonal entry is $\pi(s)$. Reward $r(s)$ is received when a transition from state $s$ takes place. Note that this reward can be stochastic as well, and the additional noise thereof can be combined with other noise terms without affecting our concentration result. For simplicity, we assume that we receive a deterministic $r(s)$. The objective is to evaluate the long term discounted reward for each state given by the ``value function''
$$V(s)=E\left[\sum_{m=0}^\infty\gamma^mr(X_m)|X_0=s\right], s\in \Scal.$$
Here $0<\gamma<1$ is the discount factor. The dynamic programming equation for evaluating the same is
$$V(s)=r(s)+\gamma\sum_{s'\in\Scal}p(s'|s)V(s'), \ s\in \Scal.$$ 
This can be written as the following vector equation
$$V=r+\gamma PV,$$
for $r=[r(1),...,r(S)]^T$ and $P=[[p(s'|s)]]_{s',s\in \Scal}\in \RR^{S\times S}$.

The state space can often be large ($S\gg1$), and to alleviate this `curse of dimensionality', $V$ is often approximated using a linear combination of $d$ linearly independent basis functions (feature vectors) $\phi_i\in\RR^S, 1\leq i\leq d$, with $S>>d\geq 1$. Also, let $\varphi(s)=[\phi_1(s),\ldots,\phi_d(s)]^T$ for $s\in\Scal$ denote the vector comprising of components corresponding to state $s$ in each feature. Thus $V(s)\approx\sum_{i=1}^dx(i)\phi_i(s)$, i.e., $V(s)\approx x^T\varphi(s)$ and $V\approx\Phi x$ where $x=[x(1),...,x(d)]^T$ and $\Phi$ is an $S\times d$ matrix whose $i$\textsuperscript{th} column is $\phi_i$. Here $x$ denotes the learnable weights for the linear function approximator. Since $\{\phi_i\}$ are linearly independent, $\Phi$ is full rank. Substituting this approximation into the dynamic programming
equation above leads to
\[
\Phi x\approx r+\gamma P\Phi x.
\]
But the RHS may not belong to the range of $\Phi$. So we use the following fixed point equation:
\begin{equation}\label{TD0-H-fixed}
    \Phi x=\Pi(r+\gamma P\Phi x)\coloneqq H(\Phi x),
\end{equation}
where $\Pi$ denotes the projection to Range($\Phi$) with respect to a suitable norm. It turns out to be convenient to take projection with respect to the weighted norm 
$\|y\|_D\coloneqq \sqrt{y^TDy}=(\sum_{s\in\Scal}\pi(s)(y(s))^2)^{1/2}$ for $y\in\RR^S$. The projection map with respect to this norm is
\begin{equation*}
    \Pi y\coloneqq\Phi(\Phi^TD\Phi)^{-1}\Phi^TDy.
\end{equation*}
The invertibility of $\Phi^TD\Phi$ is guaranteed by the fact that $\Phi$ is full rank.
Finally, the TD(0) algorithm is given by the recursion
\begin{equation}\label{TD0}
    x_{n+1}=x_n+a(n)\varphi(Y_n)\Big(r(Y_n)+\gamma\varphi(Y_{n+1})^Tx_n-\varphi(Y_n)^Tx_n\Big).
\end{equation}
Here $a(n)$ denotes the positive stepsize sequence. At the end of this section, we explain how this iteration can be expected to converge to the required fixed point from \eqref{TD0-H-fixed}.

\subsection{Assumptions}
We impose two assumptions on the algorithm. The first is about the feature vectors, and as we explain next, it does not restrict the algorithm. The second specifies the class of stepsizes $a(n)$ considered, which is standard in analysis of RL. In fact, our results hold for a broader class of step sizes than those typically required in stochastic approximation frameworks.

\begin{enumerate}
    \item For the assumption on $\Phi$, define $\Psi\coloneqq\Phi^T\sqrt{D}$ and let $\lambda_M$ be the largest singular value of $\Psi$, i.e., the square of the largest eigenvalue of $\Psi\Psi^T$ and equivalently, of $\Psi^T\Psi$. Assume that
    \begin{equation}\label{TD0_assumption}
        \lambda_M<\frac{\sqrt{2(1-\gamma)}}{(1+\gamma)}.
    \end{equation}
    Since the feature vectors can be scaled without affecting the algorithm (the weights $x(i)$ get scaled accordingly), this assumption does not restrict the algorithm. Alternatively, the stepsize can be appropriately scaled as well, i.e., $a(n)=b(n)c$, where $b(n)$ acts as the effective stepsize and $c\varphi(\cdot)$ act as the effective feature vectors which satisfy the above assumption. This assumption can be replaced with the following assumption on the basis vectors. $$\|\varphi(s)\|\leq \frac{\sqrt{2(1-\gamma)}}{1+\gamma}\;\forall s\in\Scal,$$
    i.e., the $\ell_2$ norm of each row of $\Phi$ is bounded by $\sqrt{2(1-\gamma)}/(1+\gamma)$. To see this, note that
    $$\frac{\|\Psi x\|_2}{\|x\|_2}=\frac{\|\Phi x\|_D}{\|x\|_2}\leq \frac{\|\Phi x\|_\infty}{\|x\|_2}.$$Now $\max_{x\neq \theta}\frac{\|\Phi x\|_\infty}{\|x\|_2}$ is the operator norm defined with $\ell_2$ norm for domain and $\ell_\infty$ for co-domain, which is equal to the maximum $\ell_2$ norm of a row. Hence $$\lambda_M=\max_{x\neq \theta}\frac{\|\Psi x\|_2}{\|x\|_2}\leq \max_{x\neq \theta}\frac{\|\Phi x\|_\infty}{\|x\|_2}=\max_{s\in\Scal}\|\varphi(s)\|_2.$$
    \item $\{a(n)\}$ is a sequence of non-negative stepsizes satisfying the conditions
    \begin{equation*}
     a(n) \to 0, \    \sum_na(n)=\infty.
    \end{equation*}
    and is assumed to be non-increasing, i.e., $a(n+1)\leq a(n) \ \forall \; n$. We also assume that $a(n)<1$ for all $n$. We further assume that $\frac{d_1}{n+1}\leq a(n)\leq d_3\left(\frac{1}{n+1}\right)^{d_2}, \forall\ n$, where $d_1>0$ and $0<d_2\leq 1$. Larger values of $d_1$ and $d_2$ and smaller values of $d_3$ improve the main result presented below. The role this assumption plays in our bounds will become clear later.

    Observe that we do not require the classical square-summability condition in stochastic approximation, viz., $\sum_na(n)^2<\infty$. This is because the contractive nature of our iterates (Lemma \ref{lemma:TD0-contraction}) gives us an additional handle on errors by putting less weight on past errors. A similar effect was observed in \cite{chandak}. The above assumptions on the stepsize sequence can be weakened so as to hold only after some $N>1$, without any changes in the analysis.
    
\end{enumerate}

\subsection{Formulation as a Stochastic Approximation Iteration}
We next rearrange algorithm (\ref{TD0}) to separate the martingale noise and the `Markov noise', and write it as a stochastic approximation iteration. 
\begin{eqnarray*}
x_{n+1}&=&x_n+a(n)\varphi(Y_n)\Big(r(Y_n)+\gamma\varphi(Y_{n+1})^Tx_n-\varphi(Y_n)^Tx_n\Big)\nonumber\\
&=&x_n+a(n)\bigg(F(x_n,Y_n)-x_n+M_{n+1}x_n\bigg)\nonumber\\
&=&x_n+a(n)\left(\sum_{s\in \Scal}\pi(s)F(x_n,s)-x_n\right)+\underbrace{a(n)M_{n+1}x_n}_{\tau_1}+\underbrace{a(n)\left(F(x_n,Y_n)-\sum_{s\in \Scal}\pi(s)F(x_n,s)\right)}_{\tau_2},
\end{eqnarray*}
where 
\[
F(x,Y)=\varphi(Y)r(Y)+\gamma\varphi(Y)\sum_{s'\in\Scal}p(s'|Y)\varphi(s')^Tx-\varphi(Y)\varphi(Y)^Tx+x,
\]
and 
\[
M_{n+1}=\gamma\varphi(Y_n)\Big(\varphi(Y_{n+1})^T-\sum_{s'\in\Scal}p(s'|Y_n)\varphi(s')^T\Big).
\]
Define the family  of $\sigma$-fields $\mathcal{F}_n\coloneqq\sigma(x_0,Y_m, m\leq n), \ n\geq0$.
Then $\{M_nx_{n-1}\}$ is a martingale difference sequence, with respect to $\{\mathcal{F}_n\}$, i.e., 
$$E[M_{n+1}x_n|\mathcal{F}_n]=\theta, a.s.\; \forall n,$$
where $\theta$ denotes the zero vector. The term $\tau_1$ denotes the error due to the martingale noise term and term $\tau_2$ denotes the error due to the `Markov noise' $\{Y_n\}$. 

The following lemma shows that the function $\sum_{s\in\Scal} \pi(s)F(\cdot,s)$ is a contraction. Let $\langle x,x'\rangle=x^Tx'$ and $\langle x,z\rangle_D=x^TDz$. While the contraction property of the TD(0) algorithm is well known \cite{vanRoyTsitsiklis}, we obtain an explicit expression for the contraction factor. 

\begin{lemma}\label{lemma:TD0-contraction}
    For any $x,z\in\RR^d$, 
    $$\|\sum_{s\in\Scal}\pi(s)(F(x,s)-F(z,s))\|\leq \alpha \|x-z\|,$$
    where $$\alpha=\sqrt{1-\min_{x\neq\theta}\frac{\|\Phi x\|_D^2}{\|x\|^2}\bigg(2(1-\gamma)-\lambda_M^2(1+\gamma)^2\bigg)}.$$ Moreover, $0<\alpha<1$ and hence the function $\sum_{s\in\Scal}\pi(s)F(\cdot,s)$ is a contraction.
\end{lemma}

The proof appears in Appendix \ref{Appendix-proofs}. The Banach contraction mapping theorem implies that $\sum_{s\in\Scal} \pi(s)F(\cdot,s)$ has a unique fixed point $x^*$, i.e., there exists a unique point $x^*$ such that $\sum_{s\in\Scal}\pi(s)F(x^*,s)=x^*$. We next show that the fixed point $x^*$ is the required fixed point we wish to converge to. Before that, we first observe that 
$$\sum_{s\in\Scal}\pi(s)F(x,s)=(\Phi^TDr+\gamma\Phi^TDP\Phi-\Phi^TD\Phi+I)x.$$
Then,
\begin{align*}
&\sum_{s\in\Scal}\pi(s)F(x^*,s)=(\Phi^TDr+\gamma\Phi^TDP\Phi-\Phi^TD\Phi+I)x^*=x^* \\
&\implies(\Phi^TDr+\gamma\Phi^TDP\Phi)x^*=\Phi^TD\Phi x^* \\
&\implies (\Phi^TD\Phi)^{-1}(\Phi^TDr+\gamma\Phi^TDP\Phi)x^*=x^* \\
&\implies \Phi(\Phi^TD\Phi)^{-1}\Phi^TD(r+\gamma P\Phi)x^*=\Phi x^* \\
&\implies H(\Phi x^*)=\Phi x^*.
\end{align*}
So $\Phi x^*$ is the required fixed point of (\ref{TD0-H-fixed}).

\section{Main Result}\label{sec:main-result}
Before stating the main result, we define the following two sequences. For $n\geq 0$,
\begin{eqnarray*}
    b_k(n) &=& \sum_{m=k}^na(m), \ 0\leq k\leq n<\infty, \\
   \beta_k(n) &=& \begin{cases}
      \frac{1}{k^{d_2-d_1}n^{d_1}}, & \text{if}\ d_1\leq d_2 \\
      \frac{1}{n^{d_2}}, & \text{otherwise}.
    \end{cases}
\end{eqnarray*}

Our main result is as follows:
\begin{theorem}\label{thm:main}
    There exist finite positive constants $c_1, c_2$ and $D$ such that for $0<\delta\leq1, 0<\epsilon\leq 1$, $n_0>0$ large enough to satisfy $\alpha+a(n_0)c_1<1$, and $n\geq n_0$,
    \begin{enumerate}[(a)]
        \item the inequality 
        $$\|x_m-x^*\|\leq e^{-(1-\alpha)b_{n_0}(m-1)}\epsilon+\frac{a(n_0)(c_2+c_1\epsilon)+\delta}{1-\alpha-a(n_0)c_1}, \; \forall n_0\leq m\leq n,$$
        holds with probability exceeding 
        $$1-2d\sum_{m=n_0+1}^n e^{-D\delta^2/\beta_{n_0}(m-1)}-P(\|x_{n_0}-x^*\|>\epsilon).$$
        \item In particular, 
        $$\|x_m-x^*\|\leq e^{-(1-\alpha)b_{n_0}(m-1)}\epsilon+\frac{a(n_0)(c_2+c_1\epsilon)+\delta}{1-\alpha-a(n_0)c_1}, \; \forall m\geq n_0,$$
        holds with probability exceeding 
        $$1-2d\sum_{m\geq n_0+1} e^{-D\delta^2/\beta_{n_0}(m-1n)}-P(\|x_{n_0}-x^*\|>\epsilon).$$
    \end{enumerate}
\end{theorem}
The following are some remarks about the theorem and the proof that follows:
\begin{remark}
    The assumption that $\delta\leq 1$ and $\epsilon\leq 1$ is used in the proof for Lemma \ref{lemma:bound_p_m} and has been made only for simplicity.  These can be taken as any positive values, with changes required only in the constant $D$. 
\end{remark}
\begin{remark}
    The term  $P(\|x_{n_0}-x^*\|>\epsilon)$ captures the unavoidable contribution of the initial condition at $n_0$. This can be bounded by combining moment bounds \citep{Srikant, Bhandari, Chen2} with Markov's inequality.
\end{remark}
\begin{remark}
        In \cite{Zaiwei_alltime}, an all-time bound is obtained which  goes to zero as $m\uparrow\infty$.  In our bound, the term $\frac{a(n_0)(c_2+c_1\epsilon)+\delta}{1-\alpha-a(n_0)c_1}$ remains constant as $m$ is increased. Here $\delta$ can be modified to $\delta(m)$ (similar to the treatment in Corollary 1 in \cite{chandak}). But the term $a(n_0)(c_2+c_1\epsilon)$ arises from our treatment of Markov noise using Poisson equation. Note that \cite{Zaiwei_alltime}
    do not consider the case of Markov noise, but only consider the case of additive and multiplicative noise (i.i.d.\ samples of state-action-next state triplets). We leave incorporating their ideas into our approach to get a bound decaying with $m$ for Markov noise as a future work. 
\end{remark}
\begin{remark}
    For the special case of $a(n)=\frac{d_1}{n+1}$, we combine our result with Theorem 2.1 from \citet{Chen2}, a mean square error bound, to get the following corollary. 
    \begin{corollary}\label{coro}
        Let $a(n)=d_1/(n+1)$ with sufficiently large $d_1$. Let $n_0$ be large enough to satisfy assumptions of Theorem \ref{thm:main} and \citep[Theorem 2.1]{Chen2}. Then with probability at least $1-\varepsilon_1-\varepsilon_2$, we have for all $m\geq n_0$ that
        $$\|x_m-x^*\|=\mathcal{O}\left(\frac{1}{\sqrt{n_0}}\log^{1/2}\left(\frac{1}{\varepsilon_1}\right)+\sqrt{\frac{\log(n_0)}{n_0}}\frac{1}{\sqrt{\varepsilon_2}}\left(\frac{n_0}{m}+\frac{1}{n_0}\right)\right).$$
    \end{corollary}
    The proof for this corollary has been presented at the end of Appendix \ref{Appendix-proofs}. The first term here corresponds to the term $\delta$ in Theorem \ref{thm:main}. This term has a $\sqrt{1/n_0}$ decay rate and an exponentially small tail. The second term is the contribution of the initial condition at $n_0$. We have a polynomial tail in this case, but the dependence on $m$ and $n_0$ is stronger, as the term $(\sqrt{\log(n_0)}/\sqrt{n_0})\times (n_0/m)$ decays with $m$ and the other term decays as $\log^{1/2}(n_0)n_0^{-3/2}$.

\end{remark}

\section{Proof of the Main Result}\label{sec:proof}
We present the proof of the main theorem in this section. The key martingale concentration inequality used in our proof is stated in Appendix \ref{Appendix-main} and proofs for the technical lemmas used in the proof are presented in Appendix \ref{Appendix-proofs}.
\begin{proof}[Proof of Theorem 1.]
Define $z_n$ for $n\geq n_0$ by 
    $$z_{n+1}=z_n+a(n)\left(\sum_{s}\pi(s)F(z_n,s)-z_n\right),$$
    where $z_{n_0}=x_{n_0}$. Note that $\|x_n-x^*\|\leq \|x_n-z_n\|+\|z_n-x^*\|$. To bound the second term, note that
    \begin{align*}
        z_{n+1}-x^*&= (1-a(n))(z_n-x^*)+a(n)\left(\sum_{s\in\Scal} \pi(s)F(z_n,s)-x^*\right)\\
        &=(1-a(n))(z_n-x^*)+a(n)\sum_{s\in\Scal}\pi(s)\left(F(z_n,s)-F(x^*,s)\right).
    \end{align*}
    The second equality follows from the fact that $x^*$ is a fixed point for $\sum_{s\in\Scal}\pi(s)F(\cdot,s)$. Then,
    \begin{align*}
   \|z_{n+1}-x^*\|&\leq(1-a(n))\|z_n-x^*\|+a(n)\|\sum_{s\in \Scal}\pi(s)(F(z_n,s)-F(x^*,s))\|\\
   &\leq(1-(1-\alpha)a(n))\|z_n-x^*\|.
\end{align*}
which finally gives us
\begin{equation}\label{Bound_z_x*}
    \|z_n-x^*\|\leq \prod_{k=n_0}^{n-1}(1-(1-\alpha)a(k))\|x_{n_0}-x^*\|\leq e^{-(1-\alpha)b_{n_0}(n-1)}\|x_{n_0}-x^*\|,
\end{equation}
This also implies that for all $n\geq n_0$,
\begin{equation}\label{z_n_bound}
    \|z_n\|\leq \|x_{n_0}-x^*\|+\|x^*\|.
\end{equation}
    
    Next we give a probabilistic bound on the term $\|x_n-z_n\|$. Note that
    \begin{align*}
        x_{n+1}-z_{n+1}=&(1-a(n))(x_n-z_n)+a(n)M_{n+1}x_n\\
        &+ a(n)\left(F(x_n,Y_n)-\sum_{s}\pi(s)F(z_n,s)\right) \\
        &=(1-a(n))(x_n-z_n) + \ a(n)M_{n+1}x_n \nonumber\\
        &+ \ a(n)\left(\sum_{s\in\Scal}\pi(s)(F(x_n,s)-F(z_n,s))\right)   \nonumber \\
        &+ \ a(n)\left(F(x_n,Y_n)-\sum_{s\in\Scal}\pi(s)F(x_n,s)\right). 
    \end{align*}

For $n,m\geq0$, let $\chi(n,m)=\prod_{k=m}^{n}(1-a(k))$ if $n\geq m$ and $1$ otherwise. For some $n\geq n_0$, we iterate the above for $n_0\leq m\leq n$ to obtain
\begin{align}\label{split}
    x_{m+1}-z_{m+1}=&\sum_{k=n_0}^m\chi(m,k+1)a(k)M_{k+1}x_k \nonumber\\
    &+\sum_{k=n_0}^m\chi(m,k+1)a(k)\left(\sum_{s\in \mathcal{S}}\pi(s)(F(x_k,s)-F(z_k,s))\right)\nonumber\\
    &+\sum_{k=n_0}^m\chi(m,k+1)a(k)\left(F(x_k,Y_k)-\sum_{s\in\Scal}\pi(s)F(x_k,s)\right).
\end{align}
Here we use the definition that $x_{n_0}=z_{n_0}$. We first simplify the third term above. For simplicity, we define $F(x,Y)=F_1(Y)+F_2(Y)x+x,$ where $$F_1(Y)=\varphi(Y)r(Y)\in\RR^d\;\;\;\textrm{and}\;\;\;F_2(Y)=\left(\gamma\varphi(Y)\sum_{s'\in\Scal}p(s'|Y)\varphi(s')^T-\varphi(Y)\varphi(Y)^T\right)\in\RR^{d\times d}.$$ 

We define $U:\Scal\mapsto \RR^d$ to be a solution of the Poisson equation:
\begin{equation}\label{Poisson_eqn_U}
    U(s)=F_1(s)-\sum_{s'\in\Scal}\pi(s')F_1(s')+\sum_{s'\in\Scal}p(s'|s)U(s'), \ s\in \Scal.
\end{equation}
For $s_0\in \Scal$, $\tau\coloneqq\min\{n>0:Y_n=s_0\}$ and $E_s[\cdots]=E[\cdots|Y_0=s]$, we know that 
\begin{equation}\label{above}
U'(s)=E_s\left[\sum_{m=0}^{\tau-1}(F_1(Y_m)-\sum_{s'\in \mathcal{S}}\pi(s')F_1(s'))\right], \ s\in S,
\end{equation}
is one particular solution to the Poisson equation (see, e.g., Lemma 4.2 and Theorem 4.2 of Section VI.4, pp.\ 85-91, of \citet{Borkar-Topics}). Thus $\|U'(s)\|_{\infty}\leq2\max_{s\in\Scal}\|F_1(s)\|_{\infty}E_s[\tau]$. For an irreducible Markov chain with a finite state space, $E_s[\tau]$ is finite for all $s$ and hence the solution $U'(s)$ is bounded for all $s$.  For each $\ell$,  the Poisson equation specifies $U^\ell(\cdot)$ uniquely only up to an additive constant.  Along with the additional constraint that $U(s_0)=0$ for a prescribed $s_0\in S$, the system of equations given by (\ref{Poisson_eqn_U}) has a unique solution. Henceforth $U$ refers to the unique solution of the Poisson equation with $U(s_0)=0$. 
Similarly, let $W:\Scal\mapsto\RR^{d\times d}$ be the unique solution of the Poisson equation:
\begin{equation}\label{Poisson_eqn_W}
    W(s)=F_2(s)-\sum_{s'}\pi(s')F_2(s')+\sum_{s'}p(s'|s)W(s), \ s\in S,
\end{equation}
with the additional constraint that $W(s_0)=0$ for a prescribed $s_0\in S$ as above. 

The following lemma gives a simplification of the third term in \eqref{split}, using the solutions of Poisson equation stated above. Before stating the lemma, we first define $x_m'=\sup_{n_0\leq k\leq m} \|x_m-z_m\|$. 
\begin{lemma}\label{lemma:U_W}
There exist positive constants $c_1,c_2$ such that for all $n_0\leq m\leq n$,
    \begin{align*}
        & \sum_{k=n_0}^m\chi(m,k+1)a(k)\left(F(x_k,Y_k)-\sum_{s\in\Scal}\pi(s)F(x_k,s)\right)\\
& = \ \sum_{k=n_0}^m\chi(m,k+1)a(k)\left(\widetilde{U}_{k+1}+\widetilde{W}_{k+1}x_k\right)+\mu_m(n_0),
    \end{align*}
    where $$\|\mu_m(n_0)\|\leq a(n_0)(c_2+c_1x_m'+c_1\|x_{n_0}-x^*\|).$$
    Here $\widetilde{U}_{k+1}$ and $\widetilde{W}_{k+1}x_k$ are martingale difference sequences with respect to $\mathcal{F}_k$ where $\widetilde{U}_{k+1} = U(Y_{k+1})-\sum_{s'}p(s'|Y_k)U(s')$ and $\widetilde{W}_{k+1}=W(Y_{k+1})-\sum_{s'}p(s'|Y_k)W(s')$ for $k\geq n_0$ and the zero vector, resp.\ the zero matrix, otherwise.
\end{lemma}

The proof appears in Appendix \ref{Appendix-proofs}. 
Returning to \eqref{split}, we now have
\begin{align*}
    x_{m+1}-z_{m+1}&=\sum_{k=n_0}^m\chi(m,k+1)a(k)\left(\sum_{s\in \mathcal{S}}\pi(s)(F(x_k,s)-F(z_k,s))\right)\\
    &\;\;\;+\sum_{k=n_0}^m \chi(m,k+1)a(k)\left(M_{k+1}x_k+\widetilde{W}_{k+1}x_k+\widetilde{U}_{k+1}\right)+\mu_m(n_0).
\end{align*}
Now,
\begin{align}\label{bound_on_x-z}
    \|x_{m+1}-z_{m+1}\|&\leq \left\|\sum_{k=n_0}^m \chi(m,k+1)a(k)\left(\sum_{s\in \mathcal{S}}\pi(s)(F(x_k,s)-F(z_k,s))\right)\right\|\nonumber\\
    &\;\;\;+ \ \left\|\sum_{k=n_0}^m \chi(m,k+1)a(k)\left(M_{k+1}x_k+\widetilde{W}_{k+1}x_k+\widetilde{U}_{k+1}\right)\right\| \nonumber \\
&\;\;\; + \ a(n_0)\left(c_2+c_1x_m'+c_1\|x_{n_0}-x^*\|\right)\nonumber\\
    &\leq  \alpha \sum_{k=n_0}^m \chi(m,k+1)a(k)\|x_k-z_k\|+a(n_0)\left(c_2+c_1x_m'+c_1\|x_{n_0}-x^*\|\right)\nonumber\\
    &\;\;\;+ \ \left\|\sum_{k=n_0}^m \chi(m,k+1)a(k)\left(M_{k+1}x_k+\widetilde{W}_{k+1}x_k+\widetilde{U}_{k+1}\right)\right\|.
\end{align}
For any $0<k\leq m$,
$$\chi(m,k)+\chi(m,k+1)a(k)=\chi(m,k+1),$$ and hence
$$\chi(m,n_0)+\sum_{k={n_0}}^m\chi(m,k+1)a(k)=\chi(m,m+1)=1.$$ This implies that 
\begin{equation*}
    \sum_{k=n_0}^m\chi(m,k+1)a(k) \ \leq \ 1.
\end{equation*}
Using the definition of $x'_m$, we have 
\begin{align}\label{bound_on_x'}
    x'_{m+1}\leq &\; (\alpha + a(n_0)c_1)x'_m+\left\|\sum_{k=n_0}^m \chi(m,k+1)a(k)\left(M_{k+1}x_k+\widetilde{W}_{k+1}x_k+\widetilde{U}_{k+1}\right)\right\|\nonumber\\
    &+a(n_0)(c_2+c_1\|x_{n_0}-x^*\|).
\end{align}

Next we wish to obtain a bound on the probability $$P\Big(\|x_m-x^*\|\leq \exp(-(1-\alpha)b_{n_0}(m-1))\epsilon+\Delta(n_0,\epsilon,\delta), \;\forall n_0\leq m\leq n\Big),$$
for some $\epsilon>0$ and $\delta>0$ (recall the assumption that $\alpha+a(n_0)c_1<1$). For ease of notation, here we have defined $$\Delta(n_0,\epsilon,\delta)\coloneqq\frac{a(n_0)(c_2+c_1\epsilon)+\delta}{1-\alpha-a(n_0)c_1}.$$ From \eqref{Bound_z_x*}, recall that $\|z_n-x^*\|\leq \exp(-(1-\alpha)b_{n_0}(n-1))\|x_{n_0}-x^*\| \; a.s.,$ and hence,
$$\|x_{n_0}-x^*\|\leq \epsilon \implies \|z_m-x^*\|\leq \exp(-(1-\alpha)b_{n_0}(m-1))\epsilon.$$
Also recall that $\sup_{n_0\leq m\leq n}\|x_m-z_m\|=x_n'$. Hence \begin{align*}
    &\left\{\|x_{n_0}-x^*\|\leq \epsilon\right\}\bigcap\left\{x'_n\leq \Delta(n_0,\epsilon,\delta)\right\}\\
    &\subseteq \left\{\|x_m-x^*\|\leq \exp(-(1-\alpha)b_{n_0}(m-1))\epsilon+\Delta(n_0,\epsilon,\delta), \;\forall n_0\leq m\leq n\right\}.
\end{align*}
This implies the following relation between the probabilities of the two sets.
\begin{align*}
    &P\Big(\|x_m-x^*\|\leq \exp(-(1-\alpha)b_{n_0}(m-1))\|x_{n_0}-x^*\|+\Delta(n_0,\epsilon,\delta), \;\forall n_0\leq m\leq n\Big)\\
    &\geq 1-P\left(\Big\{x'_n>\Delta(n_0,\epsilon,\delta)\Big\}\bigcup \Big\{\|x_{n_0}-x^*\|>\epsilon\Big\}\right).
\end{align*}


To compensate for the lack of an almost sure bound on the iterates $\{x_n\}$, we adapt the proof method from Proposition 34 from \cite{tao} (see Section 8 of \cite{Chung_Mart_ineq} for a detailed explanation). For this, we define $\xi=\{x_0, Y_k, k\geq0\}$ and the `bad' set $\BB_m$ as $$\BB_m=\left\{\xi\mid x_m'(\xi)>\Delta(n_0,\epsilon,\delta)\;\; \bigcup \;\;\|x_{n_0}(\xi)-x^*\|>\epsilon\right\}.$$ 
Here the notation $x_m'(\xi)$ and $x_{n_0}(\xi)$ highlights the dependence of $x_m'$ and $x_{n_0}$ on the realizations of $x_0$ and $\{Y_k\}$. Analogous notation is used for other random variables. For $\xi\notin\BB_{n-1}$ let us define $\bar{x}_{k,n-1}(\xi)=x_k(\xi)$ and $\bar{z}_{k,n-1}(\xi)=z_k(\xi)$ for all $k$. For $\xi\in\BB_{n-1}$, we define $\bar{x}_{k,n-1}(\xi)=x^*$, and $\bar{z}_{k,n-1}(\xi)=x^*$ for all $k$. Also, define $\bar{x}_{m,n-1}'(\xi)=\sup_{n_0\leq k\leq m} \|\bar{x}_{k,n-1}(\xi)-\bar{z}_{k,n-1}(\xi)\|$. Note that, $\bar{x}'_{m,n-1}=0$ when $\xi\in\BB_{n-1}$, and $\bar{x}'_{m,n-1}=x'_m\leq\Delta(n_0,\epsilon,\delta)$ when $\xi\notin\BB_{n-1}$. The intuition behind these definitions is that $\bar{x}'_{m,n-1}$ is always bounded by $\Delta(n_0,\epsilon,\delta)$ for all $m\leq n-1$. 

Note that $\xi\notin\BB_{n-1}\implies \bar{x}_{n,n-1}'(\xi)=x_n'(\xi)$, which implies
$P(\bar{x}_{n,n-1}'(\xi)\neq x_n'(\xi))\leq P(\BB_{n-1}).$ Henceforth we drop $\xi$ for ease of notation, rendering implicit the dependence of all random variables on $\xi$. Then,
\begin{align*}
    P\left(x'_n>\Delta(n_0,\epsilon,\delta)\right)&\leq P\left(x'_n>\Delta(n_0,\epsilon,\delta)\;\bigcup\;\|x_{n_0}-x^*\|>\epsilon\right)\\
    &\stackrel{(a)}{\leq} P\left(\bar{x}'_{n,n-1}>\Delta(n_0,\epsilon,\delta)\;\bigcup\;\bar{x}'_{n,n-1}\neq x'_{n}\;\bigcup\;\|x_{n_0}-x^*\|>\epsilon \right)\\
    &\stackrel{(b)}{\leq}  P\left(\bar{x}'_{n,n-1}>\Delta(n_0,\epsilon,\delta)\right)+P\left(\bar{x}_{n,n-1}'\neq x_n'\;\bigcup\;\|x_{n_0}-x^*\|>\epsilon\right)\\
    &\stackrel{(c)}{\leq} P\left(\bar{x}'_{n,n-1}>\Delta(n_0,\epsilon,\delta)\right)+ P(\BB_{n-1})\\
    &= P\left(\bar{x}'_{n,n-1}>\Delta(n_0,\epsilon,\delta)\right)+ P\left(x_{n-1}'>\Delta(n_0,\epsilon,\delta)\;\; \bigcup \;\;\|x_{n_0}-x^*\|>\epsilon\right).
\end{align*}
Inequality (a) here follows from the observation that 
$$\left\{\bar{x}'_{n,n-1}\leq\Delta(n_0,\epsilon,\delta)\right\}\bigcap \{\bar{x}'_{n,n-1}=x'_{n}\}\subseteq \left\{x'_n\leq\Delta(n_0,\epsilon,\delta)\right\},$$ which implies that
$$\left\{x'_n>\Delta(n_0,\epsilon,\delta)\right\}\subseteq\left\{\bar{x}'_{n,n-1}>\Delta(n_0,\epsilon,\delta)\right\}\bigcup \{\bar{x}'_{n,n-1}\neq x'_{n}\},$$ which gives us the required inequality. Inequality (b) follows from union bound and inequality (c) follows from the observations that $\{\|x_{n_0}-x^*\|>\epsilon\}\subseteq \BB_{n-1}$ and $\{\bar{x}_{n,n-1}'\neq x_n'\}\subseteq \BB_{n-1}$.

Now we obtain a bound for $P\left(\bar{x}'_{n,n-1}>\Delta(n_0,\epsilon,\delta)\right)$ by induction. We first note that $\bar{x}'_{n-1,n-1}$ is bounded by $\Delta(n_0,\epsilon,\delta)$ by definition. Hence $P\left(\bar{x}'_{n,n-1}>\Delta(n_0,\epsilon,\delta)\right)=P\left(\|\bar{x}_{n,n-1}-\bar{z}_{n,n-1}\|>\Delta(n_0,\epsilon,\delta)\right)$.
We first restate \eqref{bound_on_x'} for $m=n-1$.
\begin{align*}
    \|x_n-z_n\|\leq x'_n\leq &\; (\alpha + a(n_0)c_1)x'_{n-1}+a(n_0)(c_2+c_1\|x_{n_0}-x^*\|)\nonumber\\
    &+\left\|\sum_{k=n_0}^{n-1}\chi(m,k+1)a(k)\left(M_{k+1}x_k+\widetilde{W}_{k+1}x_k+\widetilde{U}_{k+1}\right)\right\|.
\end{align*}
Now, let $I\{\cdot\}$ denote the indicator function which is $1$ when $\{\cdot\}$ holds true, and zero otherwise.
\begin{align*}
    &\|\bar{x}_{n,n-1}-\bar{z}_{n,n-1}\|\\
    &=\|\bar{x}_{n,n-1}-\bar{z}_{n,n-1}\|I\{\xi_{n-1}\in\BB_{n-1}\}+\|\bar{x}_{n,n-1}-\bar{z}_{n,n-1}\|I\{\xi_{n-1}\notin\BB_{n-1}\}\\
    &\stackrel{(a)}{=}0\times I\{\xi_{n-1}\in\BB_{n-1}\}+\|x_n-z_n\|\times I\{\xi_{n-1}\notin\BB_{n-1}\}\\
    &\leq I\{\xi_{n-1}\notin\BB_{n-1}\}\times \Bigg((\alpha + a(n_0)c_1)x'_{n-1}+a(n_0)(c_2+c_1\|x_{n_0}-x^*\|)\\
    &\;\;\;\;\;\;\;\;\;\;\;\;\;\;\;\;\;\;\;\;\;\;\;\;\;\;\;\;\;\;\;\;\;\;\;\;\;\;\;\;\;\;\;\;\;\;\;\;\;\;+\left\|\sum_{k=n_0}^{n-1} \chi(m,k+1)a(k)\left(M_{k+1}x_k+\widetilde{W}_{k+1}x_k+\widetilde{U}_{k+1}\right)\right\|\Bigg)\\
    &\stackrel{(b)}{\leq} I\{\xi_{n-1}\notin\BB_{n-1}\}\times \Bigg((\alpha + a(n_0)c_1)\Delta(n_0,\epsilon,\delta)+a(n_0)(c_2+c_1\epsilon)\\
    &\;\;\;\;\;\;\;\;\;\;\;\;\;\;\;\;\;\;\;\;\;\;\;\;\;\;\;\;\;\;\;\;\;\;\;\;\;\;\;\;\;\;\;\;\;\;\;\;\;\;+\left\|\sum_{k=n_0}^{n-1}\chi(n-1,k+1)a(k)\left(M_{k+1}\bar{x}_{k,n-1}+\widetilde{W}_{k+1}\bar{x}_{k,n-1}+\widetilde{U}_{k+1}\right)\right\|\Bigg).
\end{align*}
Here inequality (a) follows from our definition of $\BB_{n-1}$ that $\|\bar{x}_{n,n-1}-\bar{z}_{n,n-1}\|=0$ when $\xi_{n-1}\in\BB_{n-1}$, and $x_n=\bar{x}_{n,n-1}$ when $\xi_{n-1}\notin\BB_{n-1}$. Inequality (b) follows from the fact that when $\xi_{n-1}\notin\BB_{n-1}$, then $x_k=\bar{x}_{k,n-1}$ for all $k$, and $\|x_{n_0}-x^*\|\leq \epsilon$. Substituting the expression for $\Delta(n_0,\epsilon,\delta)$ we obtain the following.
\begin{align*}
    \|\bar{x}_{n,n-1}-\bar{z}_{n,n-1}\|&\leq (\alpha + a(n_0)c_1)\frac{a(n_0)(c_2+c_1\epsilon)+\delta}{1-\alpha-a(n_0)c_1}+ a(n_0)(c_2+c_1\epsilon)\\
    &\;\;\;+\left\|\sum_{k=n_0}^{n-1}\chi(n-1,k+1)a(k)\left(M_{k+1}\bar{x}_{k,n-1}+\widetilde{W}_{k+1}\bar{x}_{k,n-1}+\widetilde{U}_{k+1}\right)\right\|\\
    &\leq \frac{a(n_0)(c_2+c_1\epsilon)}{1-\alpha-a(n_0)c_1}+\frac{\alpha + a(n_0)c_1}{1-\alpha-a(n_0)c_1}\delta\\
    &\;\;\;+\left\|\sum_{k=n_0}^{n-1} \chi(n-1,k+1)a(k)\left(M_{k+1}\bar{x}_{k,n-1}+\widetilde{W}_{k+1}\bar{x}_{k,n-1}+\widetilde{U}_{k+1}\right)\right\|.
\end{align*}

When 
$$\left\|\sum_{k=n_0}^{n-1} \chi(n-1,k+1)a(k)\left(M_{k+1}\bar{x}_{k,n-1}+\widetilde{W}_{k+1}\bar{x}_{k,n-1}+\widetilde{U}_{k+1}\right)\right\|\leq \delta,$$
we have $$\|\bar{x}_{n,n-1}-\bar{z}_{n,n-1}\|\leq \frac{a(n_0)(c_2+c_1\epsilon)+\delta}{1-\alpha-a(n_0)c_1}.$$ Hence,
\begin{align*}
    &P\left(\bar{x}'_{n,n-1}>\frac{a(n_0)(c_2+c_1\epsilon)+\delta}{1-\alpha-a(n_0)c_1}\right)\\
    &\leq P\left(\left\|\sum_{k=n_0}^{n-1} \chi(n-1,k+1)a(k)\left(M_{k+1}\bar{x}_{k,n-1}+\widetilde{W}_{k+1}\bar{x}_{k,n-1}+\widetilde{U}_{k+1}\right)\right\|> \delta\right).
\end{align*}
Let us denote the probability on the right side of the inequality as $p_{n-1}$. Then
$$P\left(x'_n>\frac{a(n_0)(c_2+c_1\epsilon)+\delta}{1-\alpha-a(n_0)c_1}\right)\leq p_{n-1}+P\left(x_{n-1}'>\frac{a(n_0)(c_2+c_1\epsilon)+\delta}{1-\alpha-a(n_0)c_1}\;\; \bigcup \;\;\|x_{n_0}-x^*\|>\epsilon\right).$$
Then repeating the same procedure using $\BB_{n-2}$, we obtain 
\begin{align*}
    &P\left(x_{n-1}'>\frac{a(n_0)(c_2+c_1\epsilon)+\delta}{1-\alpha-a(n_0)c_1}\;\; \bigcup \;\;\|x_{n_0}-x^*\|>\epsilon\right)\\
    &\leq
    p_{n-2}+P\left(x_{n-2}'>\frac{a(n_0)(c_2+c_1\epsilon)+\delta}{1-\alpha-a(n_0)c_1}\;\; \bigcup \;\;\|x_{n_0}-x^*\|>\epsilon\right).
\end{align*}
Iterating this for $n\geq m\geq n_0+1$, we get 
$$P\left(x'_n>\frac{a(n_0)(c_2+c_1\epsilon)+\delta}{1-\alpha-a(n_0)c_1}\right)\leq \sum_{m={n_0+1}}^n p_{m-1}+P(\|x_{n_0}-x^*\|>\epsilon).$$ The probabilities $p_m$ can be bounded using standard martingale inequalities as the terms of the martingale difference sequence are almost surely bounded. The following lemma, proved in Appendix \ref{Appendix-proofs},  gives a bound on the probabilities $p_m$:
\begin{lemma}\label{lemma:bound_p_m}
    There exists positive constant $D$ such that for $0<\epsilon\leq 1,0<\delta\leq 1$, 
    $$p_m\leq 2de^{-D\delta^2/\beta_{n_0}(m)}.$$ Recall that $d$ here denotes the dimension of the iterates $\{x_n\}$.
\end{lemma}

This completes the proof for the first part of Theorem \ref{thm:main}. 

 Let $A_n$ be the set $$\left\{\|x_m-x^*\|\leq e^{-(1-\alpha)b_{n_0}(m-1)}\epsilon+\frac{a(n_0)(c_2+c_1\epsilon)+\delta}{1-\alpha-a(n_0)c_1}, \; \forall n_0\leq m\leq n\right\}.$$ Then $\{A_n\}$ is a decreasing sequence of sets, i.e., $A_{n+1}\subseteq A_{n}$ for all $n\geq n_0$. Now let $A$ be the set $$\left\{\|x_m-x^*\|\leq e^{-(1-\alpha)b_{n_0}(m-1)}\epsilon+\frac{a(n_0)(c_2+c_1\epsilon)+\delta}{1-\alpha-a(n_0)c_1}, \; \forall m\geq n_0\right\}.$$ Then $A=\cap_{n=n_0}^\infty A_n$. Hence $P(A)=\lim_{n\uparrow\infty}P(A_n)$. This completes the proof for Theorem 1. 
 \end{proof}

\section{Conclusions}
In conclusion, we note some future directions. The concept of relaxed martingale concentration inequalities can be used to obtain bounds of the similar flavor for algorithms which suffer from similar issues. These include TD($\lambda$) and SSP Q Learning. Alternatively, similar bounds can be obtained for variants of temporal difference learning \citep{Chen2}. Another direction could be to improve the bounds in this paper to get an exponentially small tail for Markovian stochastic approximation.

\appendix
\section{Appendix A: A Martingale Inequality}\label{Appendix-main}
Let $\{M_n\}$ be a real valued martingale difference sequence with respect to an increasing family of $\sigma$-fields $\{\mathcal{F}_n\}$. Assume that there exist $\varepsilon, C > 0$ such that
$$E\left[e^{\varepsilon |M_n|}\Big|\mathcal{F}_{n-1}\right] \leq C \ \ \forall \; n \geq 1, \mbox{a.s.}$$
Let $S_n := \sum_{m=1}^n\zeta_{m,n}M_m$, where $\zeta_{m,n}, \ m \leq n,$, for each $n$, are a.s.\ bounded $\{\mathcal{F}_n\}$-previsible random variables, i.e., $\zeta_{m,n}$ is $\mathcal{F}_{m-1}$-measurable $\forall \; m \geq 1$, and $|\zeta_{m,n}| \leq A_{m,n}$ a.s.\ for some constant $A_{m,n}$, $\forall \; m, n$. Suppose
$$\sum_{m=1}^nA_{m,n} \leq \gamma_1, \ \max_{1\leq m \leq n}A_{m,n} \leq \gamma_2\omega(n),$$
for some $\gamma_i, \omega(n) > 0, \ i = 1,2; n \geq 1$. Then we have:

\begin{theorem}\label{thm:mart_ineq} There exists a constant $D > 0$ depending on $\varepsilon, C, \gamma_1, \gamma_2$ such that for $\epsilon > 0$,
\begin{eqnarray}
P\left(|S_n| > \epsilon\right) &\leq& 2e^{-\frac{D\epsilon^2}{\omega(n)}}, \ \ \mbox{if} \ \epsilon \in \left(0, \frac{C\gamma_1}{\varepsilon}\right], \label{LW1} \\
&&  2e^{-\frac{D\epsilon}{\omega(n)}},  \ \ \mbox{otherwise.} \label{LW2}
\end{eqnarray}
\end{theorem}

This is a variant of Theorem 1.1 of \citet{LW}. See \citet{Gugan}, Theorem A.1, pp.\ 21-23, for details.

\section{Appendix B: Technical Proofs}\label{Appendix-proofs}
\subsection{Proof of Lemma \ref{lemma:TD0-contraction}}
\begin{proof}
\begin{eqnarray}\label{TD0-complete-1}
    \|\sum_{s\in\Scal}\pi(s)(F(x,s)-F(z,i))\|^2&=&\|\gamma\sum_{s\in\Scal}\pi(s)\varphi(s)\sum_{s'\in\Scal}p(s'|s)\varphi(s')^T(x-z)\nonumber\\
    &&\;\;\;\;\;\;\;-\sum_{s\in\Scal}\pi(s)\varphi(s)\varphi(s)^T(x-z) + (x-z)\|^2\nonumber\\
    &=&\|(\gamma\Phi^TDP\Phi-\Phi^TD\Phi+I)(x-z)\|^2\nonumber\\
    &=&\|(\gamma\Phi^TDP\Phi-\Phi^TD\Phi)(x-z)\|^2 \nonumber\\
    &&+(x-z)^T(x-z)\nonumber\\
    &&-2(x-z)^T\Phi^TD\Phi(x-z)\nonumber\\
    &&+(x-z)^T(\gamma\Phi^TDP\Phi+\gamma\Phi^TP^TD\Phi)(x-z).
\end{eqnarray}
Now, 
\begin{eqnarray}\label{TD0-mani1}
(x-z)^T(\gamma\Phi^TDP\Phi+\gamma\Phi^TP^TD\Phi)(x-z)&=&(x-z)^T\gamma\Phi^T(DP+P^TD)\Phi(x-z)\nonumber\\
&=&\gamma \langle \Phi(x-z),P\Phi(x-z)\rangle_D\nonumber\\
&&+\gamma \langle P\Phi(x-z),\Phi(x-z)\rangle_D\nonumber\\
&\stackrel{(a)}{\leq}&2\gamma \|P\Phi(x-z)\|_D\|\Phi(x-z)\|_D\nonumber\\
&\stackrel{(b)}{\leq}&2\gamma\|\Phi(x-z)\|_D^2,
\end{eqnarray}
and 
\begin{eqnarray}\label{TD0-mani2}
2(x-z)^T\Phi^TD\Phi(x-z)&=&2\langle \Phi(x-z), \Phi(x-z)\rangle_D\nonumber\\
&=&2\|\Phi(x-z)\|_D^2.
\end{eqnarray}
Inequality (a) follows from the Cauchy-Schwarz inequality and (b) follows from the observation that $\|Py\|_D\leq\|y\|_D$, which can be proved as follows. 
$$\|Py\|^2_D=\sum_{s\in \Scal}\pi(s)\left(\sum_{s'\in \Scal} p(s'|s)y(s')\right)^2\leq \sum_{s\in \Scal}\pi(s)\sum_{s'\in \Scal} p(s'|s)y(s')^2=\sum_{s'\in \Scal}\pi(s')y(s')^2=\|y\|^2_D.$$
Here the inequality follows from Jensen's inequality.

Combining (\ref{TD0-mani1}) and (\ref{TD0-mani2}) with (\ref{TD0-complete-1}) gives us:
\begin{eqnarray}\label{TD0-complete-2}
\|\sum_{s\in\Scal}\pi(s)(F(x,s)-F(z,s))\|^2&\leq&\|x-z\|^2-2(1-\gamma)\|\Phi(x-z)\|_D^2\nonumber\\
&&+\|(\gamma\Phi^TDP\Phi-\Phi^TD\Phi)(x-z)\|^2.
\end{eqnarray}
 To analyze the last term in (\ref{TD0-complete-2}), we use the fact that the operator norm of a matrix defined as $\|M\|=\sup_{x\neq\theta}\frac{\|Mx\|}{\|x\|}$, using the Euclidean norm for vectors, is equal to the largest singular value of that matrix. Thus
\begin{align}\label{TD0-mani3}
\|(\gamma\Phi^TDP\Phi-\Phi^TD\Phi)(x-z)\|^2&=\|\Phi^T\sqrt{D}(\gamma\sqrt{D}P\Phi-\sqrt{D}\Phi)(x-z)\|^2\nonumber\\
&\leq\lambda_M^2\|(\gamma\sqrt{D}P\Phi-\sqrt{D}\Phi)(x-z)\|^2\nonumber\\
&=\lambda_M^2 \langle(\gamma P-I)\Phi(x-z),(\gamma P-I)\Phi(x-z)\rangle_D\nonumber\\
&=\lambda_M^2 \|(I-\gamma P)\Phi(x-z)\|_D^2\nonumber\\
&\leq \lambda_M^2(1+\gamma)^2\|\Phi(x-z)\|_D^2.
\end{align}

The last inequality follows from the triangle inequality. We now invoke assumption (\ref{TD0_assumption}) and combine (\ref{TD0-mani3}) with (\ref{TD0-complete-2}) as follows:
\begin{eqnarray}\label{TD0-final}
\|\sum_{s\in\Scal}\pi(s)(F(x,s)-F(z,s))\|^2&\leq&\|x-z\|^2-2(1-\gamma)\|\Phi(x-z)\|_D^2+\lambda_M^2(1+\gamma)^2\|\Phi(x-z)\|_D^2\nonumber\\
&<&\|x-z\|^2-2(1-\gamma)\|\Phi(x-z)\|_D^2\nonumber\\
&&+\left(\frac{\sqrt{2(1-\gamma)}}{1+\gamma}\right)^2(1+\gamma)^2\|\Phi(x-z)\|_D^2\nonumber\\
&=&\|x-z\|^2.
\end{eqnarray}
This gives us the required contraction property with contraction factor $\alpha$ for which an explicit expression can be obtained, using the first inequality in (\ref{TD0-final}), as
\begin{equation*}
        \alpha=\sqrt{1-\min_{x\neq\theta}\frac{\|\Phi x\|_D^2}{\|x\|^2}\bigg(2(1-\gamma)-\lambda_M^2(1+\gamma)^2\bigg)}.
\end{equation*}
Note that as the columns of $\Phi$ are linearly independent, $x\neq\theta\implies\Phi x\neq\theta$ and hence $\frac{\|\Phi x\|_D}{\|x\|}>0$ when $x\neq\theta$. Also, note that $\min_{x\neq \theta} \frac{\|\Phi x\|_D}{\|x\|}=\min_{\|x\|=1}\|\Phi x\|_D$ and hence by extreme value theorem, we have that $\min_{\|x\|=1}\|\Phi x\|_D$ is attained and is greater than $0$. Along with assumption (\ref{TD0_assumption}), this implies that $\alpha<1$. 
    \end{proof}

\subsection{Proof of Lemma \ref{lemma:U_W}}
\begin{proof}
Using definitions of $U(\cdot)$ and $W(\cdot)$ we have,
    \begin{subequations}
   \begin{align}
&\sum_{k=n_0}^m\chi(m,k+1)a(k)\left(F(x_k,Y_k)-\sum_{s\in\Scal}\pi(s)F(x_k,s)\right)\nonumber\\
&= \sum_{k=n_0}^m\chi(m,k+1)a(k)\left(U(Y_k)-\sum_{s'\in\Scal}p(s'|Y_{k})U(s')\right)\label{U-eqn}\\
&\;\;\;+\sum_{k=n_0}^m\chi(m,k+1)a(k)\left(W(Y_k)-\sum_{s'\in\Scal}p(s'|Y_{k})W(s')\right)x_k.\label{W-eqn}
\end{align} 
\end{subequations}
We first simplify \eqref{U-eqn} as follows:
\begin{subequations}
    \begin{align}
        &\sum_{k=n_0}^m\chi(m,k+1)a(k)\left(U(Y_k)-\sum_{s'\in\Scal}p(s'|Y_{k})U(s')\right)\nonumber\\
        &=\sum_{k=n_0}^m\chi(m,k+1)a(k)\left(U(Y_{k+1})-\sum_{s'\in\Scal}p(s'|Y_k)U(s')\right)\label{U-split1}\\
        &\;\;\;+\sum_{k=n_0+1}^m\left((\chi(m,k+1)a(k)-\chi(m,k)a(k-1))U(Y_k)\right)\label{U-split2}\\
        &\;\;\;+\chi(m,n_0+1)a(n_0)U(Y_{n_0})-\chi(m,m+1)a(m)U(Y_{m+1}).\label{U-split3}
    \end{align}
\end{subequations}
For \eqref{U-split1}, define $\widetilde{U}_{k+1}=U(Y_{k+1})-\sum_{s'\in\Scal}p(s'|Y_{k})U(s')$ for $k\geq n_0$ and $0$ otherwise. This is a martingale difference sequence with respect to $\{\mathcal{F}_n\}$. 

We define $U_{max}\coloneqq\max_{i\in S}\|U(i)\|$ and bound the norm of \eqref{U-split2} as follows:
\begin{align}
&\left\|\sum_{k=n_0+1}^m\left((\chi(m,k+1)a(k)-\chi(m,k)a(k-1))U(Y_k)\right)\right\|\nonumber\\
&\leq \left\|\sum_{k=n_0+1}^m((\chi(m,k+1)a(k)-\chi(m,k+1)a(k-1))U(Y_k)\right\|\nonumber\\
&\;\;\;\;\;\;\;+ \left\|\sum_{k=n_0+1}^m((\chi(m,k+1)a(k-1)-\chi(m,k)a(k-1))U(Y_k)\right\|\nonumber\\
&\leq \sum_{k=n_0+1}^m((a(k-1)-a(k))\chi(m,k+1)U_{max})+\sum_{k=n_0+1}^m((\chi(m,k+1)-\chi(m,k))a(k-1)U_{max})\nonumber\\
&\leq \sum_{k=n_0+1}^m((a(k-1)-a(k))U_{max})+\sum_{k=n_0+1}^m((\chi(m,k+1)-\chi(m,k))a(n_0)U_{max})\nonumber\\
&= (a(n_0)-a(m))U_{max} + (\chi(m,m+1)-\chi(m,n_0+1))a(n_0)U_{max}\nonumber\\
&\leq 2a(n_0)U_{max}.
\end{align}
The second and third inequality follow from $a(k-1)-a(k) \geq 0$  because  $a(k)$ is a non-increasing sequence for $k>n_0$, and $\chi(m,k+1)-\chi(m,k)$ is positive because $1\geq\chi(m,k+1)\geq\chi(m,k)$ for $m,k>n_0$, as $a(k)<1$ for $k>n_0$. Note that the norm of \eqref{U-split3} is directly bounded by $2a(n_0)U_{max}$. 

Now we simplify \eqref{W-eqn} as follows:
\begin{subequations}
    \begin{align}
        &\sum_{k=n_0}^m\chi(m,k+1)a(k)\left(W(Y_k)-\sum_{s'\in\Scal}(p(s'|Y_{k})W(s')\right)x_k\nonumber\\
        &=\sum_{k=n_0}^m\chi(m,k+1)a(k)\left(W(Y_{k+1})-\sum_{s'\in\Scal}(p(s'|Y_k)W(s')\right)x_k\label{W-split1}\\
        &\;\;\;+\sum_{k=n_0+1}^m (\chi(m,k+1)a(k)-\chi(m,k)a(k-1))W(Y_k)x_k\label{W-split2}\\
        &\;\;\;+\sum_{k=n_0+1}^m \chi(m,k)a(k-1) W(Y_k)(x_k-x_{k-1})\label{W-split3}\\
        &\;\;\;+\chi(m,n_0+1)a(n_0)W(Y_{n_0})x_{n_0}-\chi(m,m+1)a(m)W(Y_{m+1})x_m.\label{W-split4}
    \end{align}
\end{subequations}
Similar to the sequence $\widetilde{U}_{k+1}$, for \eqref{W-split1}, define $\widetilde{W}_{k+1}=W(Y_{k+1})-\sum_{s'\in\Scal}p(s'|Y_k)W(s')$ for $k\geq n_0$ and $0$ otherwise. Note that $\widetilde{W}_{k+1}x_k$ is a martingale difference sequence with respect to $\{\mathcal{F}_n\}$.

Define $W_{max}\coloneqq\max_{i\in S}\|W(i)\|$. Note that here $\|W(i)\|$ denotes the operator norm of a matrix, i.e., $\|W(i)\|=\sup_{x\neq\theta}\frac{\|W(i)x\|}{\|x\|}$, using the Euclidean norm for vectors. Similar to \eqref{U-split2}, we bound the norm of \eqref{W-split2} as follows:
\begin{align*}
    &\left\|\sum_{k=n_0+1}^m (\chi(m,k+1)a(k)-\chi(m,k)a(k-1))W(Y_k)x_k\right\|\\
    &\leq \left\|\sum_{k=n_0+1}^m (\chi(m,k+1)a(k)-\chi(m,k)a(k-1))W(Y_k)(x_k-z_k)\right\|\\
    &\;\;\; + \left\|\sum_{k=n_0+1}^m (\chi(m,k+1)a(k)-\chi(m,k)a(k-1))W(Y_k)z_k\right\|\\
    &\leq 2a(n_0)W_{max}(x_m'+\|x_{n_0}-x^*\|+\|x^*\|).
\end{align*}
The last inequality here follows from the definition of $x_m'=\sup_{n_0\leq k\leq m}\|x_m-z_m\|$ and from the bound on $\|z_n\|$ \eqref{z_n_bound}. For \eqref{W-split3}, let us first bound $\|x_k-x_{k-1}\|$.
\begin{align*}
    \|x_k-x_{k-1}\|&= a(k)\left\|\varphi(Y_{k-1})\left(r(Y_{k-1})+\gamma \varphi(Y_{k})^Tx_{k-1}-\varphi(Y_{k-1})^Tx_{k-1}\right)\right\|\\
    &\leq a(k)\left(K_1+K_2\|x_{k-1}\|\right)\\
    &\leq a(n_0)\left(K_1+K_2\left(x_m'+\|x_{n_0}-x^*\|+\|x^*\|\right)\right),
\end{align*}
for appropriate $K_1$ and $K_2$. Before simplifying \eqref{W-split3}, we first need to repeat an important simplification from our main proof. Note that for any $0<k\leq m$,
$$\chi(m,k)+\chi(m,k+1)a(k)=\chi(m,k+1),$$ and hence
$$\chi(m,n_0)+\sum_{k={n_0}}^m\chi(m,k+1)a(k)=\chi(m,m+1)=1.$$ This implies that 
\begin{equation*}
    \sum_{k=n_0}^m\chi(m,k+1)a(k) \ \leq \ 1.
\end{equation*}
We can finally bound the norm of \eqref{W-split3}:
\begin{align*}
    &\left\|\sum_{k=n_0+1}^m \chi(m,k)a(k-1) W(Y_k)(x_k-x_{k-1})\right\|\\
    &\leq \sum_{k=n_0+1}^m \chi(m,k)a(k-1) \left\|W(Y_k)(x_k-x_{k-1})\right\|\\
    &\leq \sum_{k=n_0+1}^m \chi(m,k)a(k-1) a(n_0)W_{max}\left(K_1+K_2\left(x_m'+\|x_{n_0}-x^*\|+\|x^*\|\right)\right)\\
    &\leq a(n_0) W_{max}\left(K_1+K_2\left(x_m'+\|x_{n_0}-x^*\|+\|x^*\|\right)\right).
\end{align*}
Finally the norm of \eqref{W-split4} can directly be bounded by 
\begin{align*}
    &\|\chi(m,n_0+1)a(n_0)W(Y_{n_0})x_{n_0}-\chi(m,m+1)a(m)W(Y_{m+1})x_m\|\\
    &\leq 2a(n_0)W_{max} \left(x_m'+\|x_{n_0}-x^*\|+\|x^*\|\right).
\end{align*}
Combining the bounds above  gives us
    \begin{align*}
        &\sum_{k=n_0}^m\chi(m,k+1)a(k)\left(F(x_k,Y_k)-\sum_{s\in\Scal}\pi(s)F(x_k,s)\right)\\
        &= \sum_{k=n_0}^m\chi(m,k+1)a(k)\left(\widetilde{U}_{k+1}+\widetilde{W}_{k+1}x_k\right)+\mu_m(n_0),
    \end{align*}
    where $$\|\mu_m(n_0)\|\leq 4a(n_0)U_{max}+a(n_0)W_{max}\left(K_1+(4+K_2)\left(x_m'+\|x_{n_0}-x^*\|+\|x^*\|\right)\right).$$

Define constants $c_1\coloneqq W_{max}(4+K_2)$ and $c_2\coloneqq4U_{max}+K_1W_{max}+c_1\|x^*\|$. This completes the proof for Lemma \ref{lemma:U_W}.
\end{proof}
\subsection{Proof of Lemma \ref{lemma:bound_p_m}}
\begin{proof}
   We first note that, for $n_0\leq k\leq m$, $\|\bar{x}_{k,m}\|\leq \|\bar{x}_{k,m}-\bar{z}_{k,m}\|+\|\bar{z}_{k,m}\|\leq \bar{x}'_{m,m}+\|\bar{z}_{k,m}\|$. The following follow from the definition of $\BB_m$. If $\xi\in\BB_m$, $\bar{x}'_{m,m}(\xi)=0$ and if $\xi\notin\BB_m$, $\bar{x}'_{m,m}(\xi)=x_m'(\xi)\leq \frac{a(n_0)(c_2+c_1\epsilon)+\delta}{1-\alpha-a(n_0)c_1}$. Hence
    $$\bar{x}'_{m,m}\leq \frac{a(n_0)(c_2+c_1\epsilon)+\delta}{1-\alpha-a(n_0)c_1}.$$ Using \eqref{z_n_bound}, we have $\|\bar{z}_{k,m}\|\leq\epsilon+\|x^*\|$. Under the condition that $\epsilon\leq 1$ and $\delta\leq1$, we have $$\|\bar{x}_{k,m}\|\leq 1+\|x^*\|+\frac{a(n_0)(c_2+c_1)+1}{1-\alpha-a(n_0)c_1}.$$ Let $v^{(\ell)}$ denote the $\ell$\textsuperscript{th} component of a vector $v$. Then 
    \begin{align*}
    \Gamma_m\coloneqq&\left\|\sum_{k=n_0}^{m} \chi(m,k+1)a(k)\left(M_{k+1}\bar{x}_{k,m}+\widetilde{W}_{k+1}\bar{x}_{k,m}+\widetilde{U}_{k+1}\right)\right\|\\
    &\leq \sqrt{d}\max_{1\leq\ell\leq d}\left|\sum_{k=n_0}^{m} \chi(m,k+1)a(k)\left(M_{k+1}\bar{x}_{k,m}+\widetilde{W}_{k+1}\bar{x}_{k,m}+\widetilde{U}_{k+1}\right)^{(\ell)}\right|.
    \end{align*}
    Recall that $d$ here is the dimension of our iterates $\{x_n\}$. We apply Theorem \ref{thm:mart_ineq} from Appendix \ref{Appendix-main} componentwise. For this, first note that $$\left(M_{k+1}\bar{x}_{k,m}+\widetilde{W}_{k+1}\bar{x}_{k,m}+\widetilde{U}_{k+1}\right)^{(\ell)}\leq c_3\left(2+\|x^*\|+\frac{a(n_0)(c_2+c_1)+1}{1-\alpha-a(n_0)c_1}\right),$$
    where $c_3=\max\{M_{max}+2W_{max},2U_{max}\}$. In the theorem statement, let
    $$C=\sqrt{d}c_3\left(2+\|x^*\|+\frac{a(n_0)(c_2+c_1)+1}{1-\alpha-a(n_0)c_1}\right), \zeta_{k,m}=\chi(m,k+1)a(k), \varepsilon=1, \gamma_1=1.$$
    Next, we choose suitable $\gamma_2$ and $\omega(m)$ such that 
    $\max_{n_0\leq k\leq m} \zeta_{k,m}\leq \gamma_2\omega(m)$. 
    For this, we use our assumption that $\frac{d_1}{n+1}\leq a(n)\leq d_3\left(\frac{1}{n+1}\right)^{d_2}, \forall\ n\geq n_0$, to obtain:
\begin{align*}
    &\chi(m,k+1)=\prod_{i=k+1}^{m}(1-a(i))\leq \exp\left(-\sum_{i=k+1}^m a(i)\right)\leq \exp \left(-\sum_{i=k+1}^m \frac{d_1}{i+1}\right) \\
    &\;\;\;\;\;\;\;\;\;\;\;\;\;\;\;\;\;\;\;\;\;\;\;\;\;\;\leq \exp\left(-\int_{k+1}^{m+1} \frac{d_1}{y+1}dy\right)\leq \exp\left(d_1(\log(k+2)-\log(m+2))\right) \\
    &\;\;\;\;\;\;\;\;\;\;\;\;\;\;\;\;\;\;\;\;\;\;\;\;\;\; =\left(\frac{k+2}{m+2}\right)^{d_1}\\
    &\implies \max_{n_0\leq k\leq m} a(k)\chi(m,k+1)\leq \max_{n_0\leq k\leq m} d_3\left(\frac{1}{k}\right)^{d_2}\left(\frac{k+2}{m+2}\right)^{d_1} \leq \max_{n_0\leq k\leq m}d_3\left(\frac{1}{k}\right)^{d_2}\left(\frac{2k}{m+2}\right)^{d_1}.
\end{align*}
From the last inequality, $\gamma_2=d_32^{d_1}$ and $\omega(m)=\beta_{n_0}(m)$ satisfy the required conditions.
Then there exists a constant $D>0$, such that for $n_0<m$ and $\delta\in(0,C]$, we have $$P(\Gamma_m\geq \delta)\leq 2de^{-D\delta^2/\beta_{n_0}(m)},$$
and for $\delta>C$, 
$$P(\Gamma_m\geq \delta)\leq 2de^{-D\delta/\beta_{n_0}(m)}.$$
The factor $d$ comes from the application of union bound to bound the maximum over all components. Under the assumption that $\delta\leq 1$, we have that $e^{-D\delta^2/\beta_{n_0}(m)}\geq e^{-D\delta/\beta_{n_0}(m)}$ and hence $P(\Gamma_m\geq \delta)\leq 2de^{-D\delta^2/\beta_{n_0}(m)}$.
\end{proof}

\subsection{Proof of Corollary \ref{coro}}
\newcommand{\sfrak}{\mathfrak{s}}
To show Corollary \ref{coro}, we first obtain values of $\delta$ and $\epsilon$ such that the probability in Theorem \ref{thm:main} is $1-\varepsilon_1-\varepsilon_2$. We use $\sfrak_i, i=1,2,\ldots$ to denote different constants throughout this proof. For $a(n)=d_1/(n+1)$ with a sufficiently large $d_1$, we have $\beta_{n_0}(m)\leq 1/m$. This implies that $$\sum_{m\geq n_0+1}\exp(-D\delta^2/\beta_{n_0}(m))\leq\sum_{m\geq n_0+1}\exp(-D\delta^2m)\leq \sfrak_1\exp(-D\delta^2n_0).$$ 
Let $\varepsilon_1/(2d)=\sfrak_1\exp(-D\delta^2n_0)$, which gives us $\delta=\sfrak_2n_0^{-1/2}\log^{1/2}(\sfrak_3/\varepsilon_1)$ for appropriate constants $\sfrak_2$ and $\sfrak_3$. This choice of $\delta$ gives us $$2d\sum_{m\geq n_0+1}\exp(-D\delta^2/\beta_{n_0}(m))\leq \varepsilon_1.$$

 Let $\epsilon=\sqrt{E[\|x_{n_0}-x^*\|^2]}/\sqrt{\varepsilon_2}$, which implies that 
$$P(\|x_{n_0}-x^*\|> \epsilon)=P\left(\|x_{n_0}-x^*\|^2> \frac{E[\|x_{n_0}-x^*\|^2]}{\varepsilon_2}\right)\leq \varepsilon_2.$$
Here the last inequality follows from Markov's inequality. Being linear contractive SA with an aperiodic irreducible Markov chain, our formulation satisfies the assumptions for Theorem 2.1 from \cite{Chen2}. To apply their result, we note the corresponding mapping between constants: the norm $\|\cdot\|_c$ is the Euclidean norm in our case, $h$ is $1$, $\varphi_2$ is $1-\alpha$ in our case and their $\alpha$ is $d_1$ in our case. For $d_1>1/(1-\alpha)$ and $n_0$ sufficiently large to satisfy the condition for Theorem 2.1 (2), we can use their result Theorem 2.1 (2)(b)(iii) to obtain the following mean square bound.
\begin{align*}
    E\left[\|x_{n_0}-x^*\|^2\right]&\leq \sfrak_4\left(\frac{1}{n_0+1}\right)^{(1-\alpha)d_1}+\sfrak_5\frac{\log(n_0+1)}{n_0+1}\leq \sfrak_6\frac{\log(n_0+1)}{n_0+1}. 
\end{align*}

Substituting the values of $\delta$ and $\epsilon$ in our bound, we get with probability greater than $1-\varepsilon_1-\varepsilon_2$, 
\begin{align*}
    \|x_m-x^*\|&\leq e^{-(1-\alpha)b_{n_0}(m-1)}\frac{\sqrt{E[\|x_{n_0}-x^*\|^2]}}{\sqrt{\varepsilon_2}}\\
    &\;\;+\sfrak_7\left(\frac{c_2d_1}{n_0+1}+\frac{c_1d_1}{n_0+1}\frac{\sqrt{E[\|x_{n_0}-x^*\|^2]}}{\sqrt{\varepsilon_2}}+\sfrak_2n_0^{-1/2}\log^{1/2}(\sfrak_3/\varepsilon_1)\right)\\
    &\leq e^{-(1-\alpha)b_{n_0}(m-1)}\frac{\sfrak_6}{\sqrt{\varepsilon_2}}\sqrt{\frac{\log(n_0+1)}{n_0+1}}\\
    &+\sfrak_7\left(\frac{c_2d_1}{n_0+1}+\frac{c_1d_1}{n_0+1}\frac{\sfrak_6}{\sqrt{\varepsilon_2}}\sqrt{\frac{\log(n_0+1)}{n_0+1}}+\sfrak_2n_0^{-1/2}\log^{1/2}(\sfrak_3/\varepsilon_1)\right),
\end{align*}
for all $m\geq n_0$. Now 
\begin{align*}
    &\exp(-(1-\alpha)b_{n_0}(m-1))\\
    &\leq \exp\left(-\sum_{i=n_0}^{m-1} (1-\alpha)a(i)\right)\leq \exp \left(-\sum_{i=n_0}^{m-1} \frac{(1-\alpha)d_1}{i+1}\right) \\
    &\leq \exp\left(-\int_{n_0}^m \frac{(1-\alpha)d_1}{y+1}dy\right)\leq \exp\left((1-\alpha)d_1(\log(k+1)-\log(m+1))\right) \\
    & =\left(\frac{n_0+1}{m+1}\right)^{d_1(1-\alpha)}\leq \frac{n_0+1}{m+1}.
\end{align*}
Here the final inequality follows from the assumption that $(1-\alpha)d_1>1$. Hence we get that for sufficiently large $n_0$, the following holds with probability $1-\varepsilon_1-\varepsilon_2$ for all $m\geq n_0$.
$$\|x_m-x^*\|=\mathcal{O}\left(\frac{1}{\sqrt{n_0}}\log^{1/2}\left(\frac{1}{\varepsilon_1}\right)+\sqrt{\frac{\log(n_0)}{n_0}}\frac{1}{\sqrt{\varepsilon_2}}\left(\frac{n_0}{m}+\frac{1}{n_0}\right)\right).$$

\section*{Acknowledgements}
The work of VSB was supported in part by S.\ S.\ Bhatnagar Fellowship from the Government of India.

\end{document}